\documentclass[11pt]{article} 

\usepackage{fullpage} 

\usepackage{float}
\restylefloat{table}

\usepackage[sort,numbers]{natbib}

\usepackage[utf8]{inputenc} 
\usepackage[T1]{fontenc}    
\usepackage{url}            
\usepackage{booktabs}       
\usepackage{amsfonts}       
\usepackage{nicefrac}       
\usepackage{microtype}      

\usepackage[utf8]{inputenc} 
\usepackage[T1]{fontenc}    
\usepackage{url}            
\usepackage{booktabs}       
\usepackage{amsfonts}       
\usepackage{nicefrac}       
\usepackage{microtype}      

\usepackage{color}
\usepackage[usenames,dvipsnames,svgnames,table]{xcolor}
\usepackage[colorlinks=true, linkcolor=red, urlcolor=blue, citecolor=blue]{hyperref}

\usepackage{algorithm}
\usepackage{algorithmic}
\usepackage{comment}
\usepackage{mathtools}
\usepackage{caption}
\usepackage{float}
\usepackage{graphicx}
\usepackage{subfigure}
\usepackage{multirow}
\usepackage{mathrsfs}
\usepackage{newfloat}
\usepackage{relsize}
\usepackage{dsfont}
\usepackage{booktabs,siunitx,caption}
\usepackage{enumitem}

\usepackage[framemethod=TikZ]{mdframed}
\newcounter{Frame}

\newcommand{\note}[1]{
	\noindent~\\
	\vspace{0.25cm}
	\fcolorbox{red}{orange}{\parbox{0.9\textwidth}{#1}}
	\vspace{0.25cm}
}

\newcommand{\notecw}[1]{{\note{CW: #1}}}

\let\citep\cite
\let\citet\cite

\title{AutoML for Climate Change: A Call to Action}

\author{
Renbo Tu$^{1}$%
\thanks{Work done while first author was part-time at Abacus.AI. Correspondence to: Colin White <\url{colin@abacus.ai}>.} ,
Nicholas Roberts$^{2}$, Vishak Prasad$^{3}$, Sibasis Nayak$^{3}$, Paarth Jain$^{3}$, \\ 
Frederic Sala$^{2}$, Ganesh Ramakrishnan$^{3}$, Ameet Talwalkar$^{4}$,
Willie Neiswanger$^{5}$, Colin White$^{6}$
    \vspace*{2mm} \\
    $^1$University of Toronto,
    $^2$University of Wisconsin,
    $^3$IIT Bombay,
    \vspace*{.5mm} \\
    $^4$Carnegie Mellon University,
    $^5$Stanford University
    $^6$Abacus.AI
  }

\date{}

\begin{document}

\maketitle

\begin{abstract}
The challenge that climate change poses to humanity has spurred a rapidly developing field of artificial intelligence research focused on climate change applications. 
The climate change AI (CCAI) community works on a diverse, challenging set of problems which often involve physics-constrained ML or heterogeneous spatiotemporal data.
It would be desirable to use automated machine learning (AutoML) techniques to automatically find high-performing architectures and hyperparameters for a given dataset.
In this work, we benchmark popular AutoML libraries on three high-leverage CCAI applications: climate modeling, wind power forecasting, and catalyst discovery. 
We find that out-of-the-box AutoML libraries currently fail to meaningfully surpass the performance of human-designed CCAI models.
However, we also identify a few key weaknesses, which stem from the fact that most AutoML techniques are tailored to computer vision and NLP applications. For example, while dozens of search spaces have been designed for image and language data, none have been designed for spatiotemporal data.
Addressing these key weaknesses can lead to the discovery of novel architectures that yield substantial performance gains across numerous CCAI applications.
Therefore, we present a call to action to the AutoML community, since there are a number of concrete, promising directions for future work in the space of AutoML for CCAI.
We release our code and a list of resources at
\url{https://github.com/climate-change-automl/climate-change-automl}.
\end{abstract}

\section{Introduction} \label{sec:introduction}

There is an increasing body of evidence which shows that climate change is one of the biggest threats facing humanity today~\citep{nakicenovic2000emissions, change2018global, archer2010climate, romm2022climate}.
Taking action towards climate change must come in many forms, such as reducing greenhouse gases and facilitating the adaption of renewable energy.
A rapidly developing area of artificial intelligence research, climate change AI (CCAI), is focused on applications to mitigate the effects of climate change~\citep{rolnick2022tackling, donti2021machine, kaack2022aligning}.

On the other hand, the automated machine learning (AutoML) community has been focused on designing efficient algorithms for problems such as hyperparameter optimization (HPO) and neural architecture search (NAS)~\citep{automl}. 
In general, the goal of AutoML is to develop algorithms that automate the process of designing architectures and tuning hyperparameters for a given dataset.
Although AutoML would seemingly be most useful on understudied datasets where there is less human intuition~\citep{roberts2021rethinking, nasbench360}, most AutoML techniques, whether implicitly or explicitly, are tailored to CV and NLP tasks.
Furthermore, a few recent works show that state-of-the-art AutoML techniques for common CV-based tasks do not transfer to other non-CV tasks~\citep{nasbenchsuite,nasbench360}.
A natural question is therefore, \emph{are AutoML techniques beneficial for high-impact CCAI applications?}

\begin{figure}
    \centering
    \includegraphics[width=\textwidth]{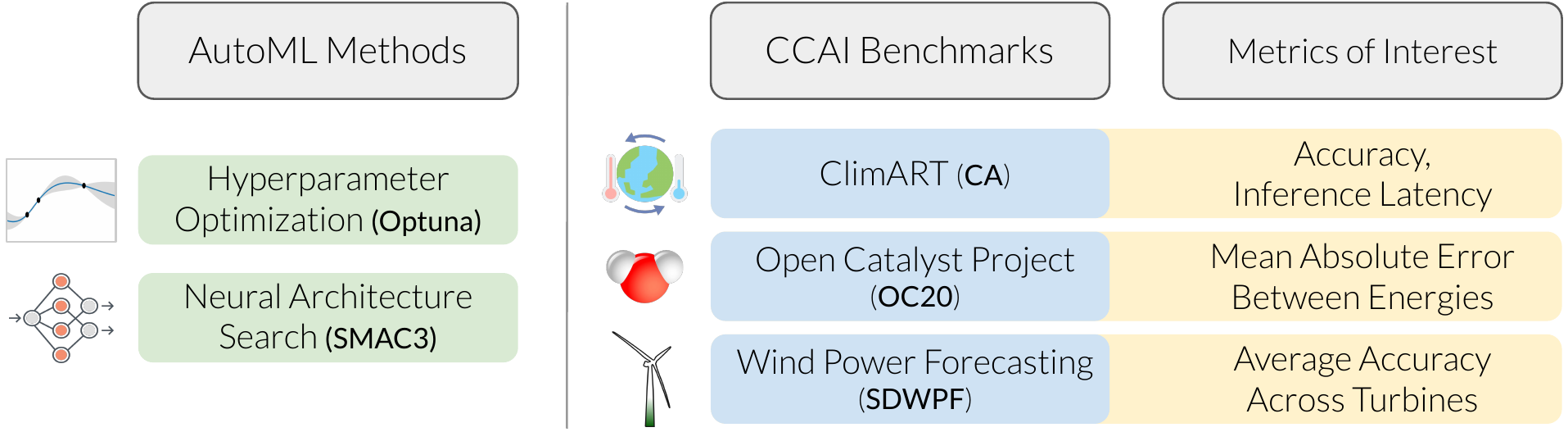}
    \vspace{-3mm}
    \caption{
    \small
    Overview of the main components of our study.}
    \label{fig:overview}
\end{figure}

In this work, we benchmark popular AutoML libraries on three high-leverage CCAI tasks: climate modeling, wind power forecasting, and catalyst discovery (see Fig.~\ref{fig:overview}).

Across several experiments, we are unable to show that out-of-the-box AutoML techniques meaningfully surpass the performance of human-designed models.
At the same time, we identify concrete weaknesses stemming from the fact that AutoML techniques have not been designed for common CCAI themes such as spatiotemporal data or physics-constrained ML.
For example, designing a search space which interpolates among MLPs, CNNs, GNNs, and GCNs (all of which have been used for climate modeling \citep{cachay2021climart,cachay2021world,liu2020radnet,pal2019using}) would allow NAS algorithms to discover novel combinations of existing architecture components, potentially leading to substantial performance gains across several spatiotemporal forecasting applications.
Therefore, we give a call to action to the AutoML community, with the aim of leveraging the full power of AutoML on challenging, high-impact CCAI tasks. 

\vspace{-1.5mm}
\paragraph{Related work.}
In recent years, several techniques have been developed for atmospheric radiative transfer \citep{cachay2021climart, brenowitz2018prognostic, rasp2018deep, yuval2020stable}, wind power forecasting \citep{zhou2022sdwpf, deng2020wind}, catalyst prediction \citep{kolluru2022open, tran2022open, das2022open}, and many more areas \citep{jain2020review, zhu2022meter, irvin2020forestnet}.
For a survey of machine learning tasks in the climate change space, see \citep{rolnick2022tackling}.

HPO \citep{feurer2019hyperparameter} and NAS \citep{nas-survey} are two popular areas of AutoML \citep{automl}.
Recently, Tu \textit{et al.}\ introduced NAS-Bench-360 \citep{nasbench360}, a benchmark suite to evaluate NAS methods on a diverse set of understudied tasks, in order to help move the field of NAS away from its emphasis on CV and NLP.
They showed that current state-of-the-art NAS methods do not perform well on diverse tasks.
Another recent work similarly showed that the best techniques and hyperparameters on CV-based tasks do not transfer to more diverse tasks \citep{nasbenchsuite}.
However, for both of these works, the analyses used a few fixed search spaces rather than identifying models hand-designed specifically for each task.

\vspace{-1mm}
\section{Methodology} \label{sec:method}
\vspace{-1mm}
In this section, we describe our methodology, driven by the following two research questions:
\begin{itemize}[topsep=0pt, itemsep=2pt, parsep=0pt, leftmargin=10mm]
    \item \textbf{RQ 1:} Can current out-of-the-box AutoML techniques substantially improve performance compared to human-designed models in high-leverage climate change AI applications?
    \item \textbf{RQ 2:} If not, then what are the key limitations and weaknesses of existing techniques?
\end{itemize}

In order to answer \textbf{RQ 1}, we select datasets which \emph{(1)} correspond to impactful directions in climate change research, and \emph{(2)} have existing strong human-designed baselines.
For example, we choose datasets which were recently featured in large competitions, with top solutions now open-source.
We describe the details of each dataset in Section \ref{sec:experiments}.

%
For each of the datasets we choose, 
we first find open-source high-performing human-designed models.
Then we run Optuna \citep{akiba2019optuna} or SMAC3 \citep{smac3}, two of the most widely-used AutoML libraries today, using top human-designed models as the base.
We compare the resulting searched models to top human-designed models.

In order to answer \textbf{RQ 2}, we check for general weaknesses in AutoML techniques applied to CCAI tasks, which can be overcome with future work.
For example, we look at whether the AutoML techniques are limited due to being implicitly tailored to CV tasks.

\section{Experiments and Discussion} \label{sec:experiments}

In this section, for three CCAI tasks, we give a brief description of the task, dataset, and our AutoML experiments.
Then, in Section \ref{subsec:discussion}, we use our experiments to answer \textbf{RQ 1} and \textbf{RQ 2}.

\subsection{Experimental Setup} \label{subsec:experiments}

\vspace{-1mm}
\paragraph{Atmospheric Radiative Transfer.}
Numerical weather prediction models, as well as global and regional climate models, 
give crucial information to policymakers and the public about the impact of changes in the Earth's climate.
The bottleneck is atmospheric radiative transfer (ART) calculations, which are used to compute the heating rate of any given layer of the atmosphere.
While ART has historically been calculated using computationally intensive physics simulations, researchers have recently used neural networks to substantially reduce the computational bottleneck, enabling ART to be run at finer resolutions and obtaining better overall predictions.

We use the ClimART dataset \citep{cachay2021climart} from the NeurIPS Datasets and Benchmarks Track 2021. It consists of global snapshots of the atmosphere across a discretization of latitude, longitude, atmospheric height, and time from 1979 to 2014.
Each datapoint contains measurements of temperature, water vapor, and aerosols.
Prior work has tested MLPs, CNNs, GNNs, and GCNs as baselines \citep{cachay2021climart}.


We run HPO on the CNN baseline from Cachay \textit{et al.}\ \citep{cachay2021climart} using the Optuna library \citep{akiba2019optuna}.
The CNN model is chosen because it had the lowest RMSE and second-lowest latency out of all five baselines from Cachay \textit{et al.}
We tune learning rate, weight decay, dropout, and batch size.
%
We also run NAS using SMAC3 \citep{smac3}.
We set a categorical hyperparameter to choose among MLP, CNN, GNN, GCN, and L-GCN \citep{cachay2021world} while also tuning learning rate and batch size.
See Appendix \ref{app:climart} for more details of the dataset and experiments.



\vspace{-1mm}
\paragraph{Wind Power Forecasting.}
Wind power is one of the leading renewable energy types, since it is cheap, efficient, and harmless to the environment \citep{foley2012current, archer2005evaluation, sadorsky2021wind}.
The only major downside in wind power is its unreliablility: changes in wind speed and direction make the energy gained from wind power inconsistent.
In order to keep the balance of energy generation and consumption on the power grid, other sources of energy must be added on short notice when wind power is down, which is not always possible (for example, coal plants take at least 6 hours to start up) \citep{hanifi2020critical}.
Therefore, forecasting wind power is an important problem that must be solved to facilitate greater adoption of wind power.

We use the SDWPF (Spatial Dynamic Wind Power Forecasting) dataset, which was recently featured in a KDD Cup 2022 competition that included 2490 participants \citep{zhou2022sdwpf}.
This is by far the largest wind power forecasting dataset, consisting of data from 134 wind turbines across 12 months.
The features consist of external features such as wind speed, wind direction, and temperature, and turbine features such as pitch angle of the blades, operating status, relative location, and elevation.
%

We use a BERT-based model 
\citep{tan2022application}, and a GRU+LGBoost model 
\citep{lin2022kdd}, which placed 3rd and 7th in the competition out of 2490, respectively (and are 1st and 3rd among open-source models, respectively).
We run HPO using Optuna for both the BERT-based model and the GRU+LGBoost model, and we also run NAS on the BERT-based model.
For additional details, see Appendix \ref{app:sdwpf}.

\vspace{-1mm}
\paragraph{Open Catalyst Project.}
Discovering new catalysts is key to cost-effective chemical reactions to address the problem of energy storage, which is necessitated by the intermittency of power generation from growing renewable sources, such as wind and solar. 
Catalyst discovery is also important for more efficient production of ammonia fertilizer, which currently makes up 1\% of the world's CO$_2$ emissions \citep{hockstad2018inventory}.
Modern methods for catalyst design use a simulation via density functional theory (DFT), which can be approximated with deep learning.
Specifically, given a set of atomic positions for the reactants and catalyst, the energy of the structure can be predicted.

We use the Open Catalyst 2020 (OC20) dataset \citep{chanussot2021open}, which was featured in a NeurIPS 2021 competition \citep{das2022open}.
Each datapoint is one reaction, where the features consist of the initial starting positions of the atoms, and the label consists of the energy needed to drive the reaction.
In our experiments, we use a downsampled version of the OC20 IS2RE out-of-distribution adsorbates task, using 59\,904 examples.

We use Graphormer \citep{shi2022benchmarking}, the winning solution from the NeurIPS 2021 Open Catalyst Challenge, developed by a team at Microsoft.
We run Optuna on the learning rate, number of warmup steps, number of layers, attention heads, and blocks. 
See Appendix \ref{app:oc20} for additional details.



\begin{table}[t]
\centering
\resizebox{.99\linewidth}{!}{%
\centering
\begin{tabular}{l|l|l|l|c|c|c|c}
\hline
Dataset & Type & Base Model & Metric & Perf. human & Perf. AutoML & improv.\ \% & search time \\ \hline
ClimART & NAS       & Various &  RMSE (W/m$^2$)   &        1.829                       &      1.669                          &          8.7\%           &      12 GPU hrs      \\ 
ClimART & HPO      & CNN        &  RMSE (W/m$^2$) &   1.829                          & 1.538                                &     15.9\%                &    54 GPU hrs        \\
\hline
SDWPF & NAS       & BERT-based    &  RMSE+MAE (kW)   &  45.246                        &          45.178                      &          0.15\%           &      26 GPU hrs       \\ 
SDWPF & HPO       & BERT-based    &   RMSE+MAE (kW)   &   45.246                         &     
45.329                      &         -0.08\%$^*$            &      42 GPU hrs      \\ 
SDWPF & HPO       & GRU+GBDT    & RMSE+MAE (kW)   & 45.074                              &          45.074                      &         0\%            &       50 GPU hrs      \\ 
\hline
OC20 & HPO       & Graphormer    &  MAE (eV) &   0.399                          &    0.396                            &           0.65\%          &   24 GPU hrs         \\
\hline
\end{tabular}
}
\caption{Empirical comparison between human-designed models and AutoML searched models. 
In the `Perf.\ AutoML' column, we report the test set performance of the model with the best validation set performance during the AutoML search ($^*$ which may be worse than the original model, if the validation set performance is higher but the test set performance is lower).
} \label{tab:results}
\end{table}

\subsection{Results and Discussion} \label{subsec:discussion}

See Table \ref{tab:results} for results and percentage improvement by running AutoML.
Despite running for 12-50 hours on each task, the AutoML techniques did not meaningfully improve performance compared to the best human-designed model, with the exception of ClimART. 
However, for ClimART, we were unable to reproduce the originally reported RMSE  of the CNN model \citep{cachay2021climart} with the default parameters, and so the AutoML performance is compared to our own (worse) evaluation of the default model.
Overall, although our experiments are not comprehensive, we find no indication that \textbf{RQ 1} is true; in other words, out-of-the-box AutoML techniques currently may not be able to substantially improve upon human-designed CCAI models.
We emphasize that our experiments were aimed specifically at evaluating AutoML methods \emph{out-of-the-box}. For a discussion of limitations, see Appendix \ref{app:experiments}.

We find that a key weakness of current AutoML methods is that the search spaces are designed for common tasks such as CV and NLP.
For example, ClimART could benefit from search spaces that interpolate among MLPs, CNNs, GNNs, and GCNs, which do not currently exist.
In general, many CCAI applications would benefit from search spaces designed specifically to handle spatiotemporal forecasting tasks, both two-dimensional \citep{zhou2022sdwpf,zhu2022meter,weather4cast,rohde2013berkeley} and three-dimensional \citep{cachay2021climart,liu2020radnet}.
Furthermore, many CCAI applications have physics constraints in some form \citep{cachay2021climart,das2022open,drgovna2021physics,drgona2020physics}. 
For example, the predictions for ART and catalyst discovery are constrained by physics laws.
Architectures which incorporate physics constraints, and loss terms with several hyperparameters, are two common methods for handling physics constraints \citep{kashinath2021physics}, and using AutoML to search for the best architecture and loss function is a promising area for future work.
Therefore, our answer to \textbf{RQ 2} is that search spaces are currently focused on CV tasks, and designing search spaces for spatiotemporal forecasting and physics constraints would be particularly beneficial across CCAI applications. 

\section{Conclusions and Future Work} \label{sec:conclusion}

In this work, we benchmarked popular AutoML libraries on datasets for climate modeling, wind power forecasting, and catalyst discovery, and we were unable to show that out-of-the-box AutoML libraries substantially improve over human-designed models.
%

There are many concrete, promising avenues for future work.
First and foremost, designing search spaces for spatiotemporal data and physics constrained ML, as mentioned in Section \ref{subsec:discussion},
could allow NAS algorithms to discover novel combinations of existing architecture components, potentially leading to substantial performance gains across several spatiotemporal forecasting applications.
Next, while our work focused on HPO and NAS, there are still many other sub-areas of AutoML, such as data augmentation, data preprocessing, and continuous monitoring and maintenance of deployed models.
Finally, while our work focused on three high-impact datasets, there are many other CCAI applications for which AutoML could be tested, such as model predictive control for buildings \citep{drgovna2020all, drgovna2018approximate} and optimal power flow \citep{fioretto2020predicting}.

\section{Broader Impact} \label{sec:impact}


Our goal in this work is to give evidence that current out-of-the-box AutoML techniques do not perform sufficiently on high-impact CCAI applications, and then
give a call to action to the AutoML community by identifying several concrete areas for future work.
The successes would be for \emph{(1)} AutoML researchers to design and test their methods on CCAI tasks, and \emph{(2)} CCAI practitioners to use (future) AutoML tools to make progress in their respective domains.

Although automated machine learning is a powerful tool to make progress on climate change problems, the large carbon footprint and financial cost of training machine learning models must be weighed \citep{schwartz2020green, tornede2021towards}.
While we strongly believe that the AutoML community heeding our call to action will have a net positive impact on society, we urge AutoML researchers to conduct research in a responsible and climate-conscious manner, using the suggestions laid out by Tornede \textit{et al.}\ \citep{tornede2021towards}.


\section*{Acknowledgments}
The authors thank Priya Donti and Ján Drgoňa for their help with this project.
This work was supported in part by the
NSF (1651565), AFOSR (FA95501910024), ARO (W911NF-21-1-0125), CZ Biohub, Sloan Fellowship,
National Science Foundation grants IIS1705121, IIS1838017, IIS2046613, IIS2112471, funding from Meta, Morgan Stanley, Amazon, Google,
National Science Foundation grants CCF2106707, and funding from Wisconsin Alumni Research Foundation (WARF). 
Any opinions, findings and conclusions or recommendations expressed in this material are those of the author(s) and do not necessarily reflect the views of any of these funding agencies.


\bibliography{main}
\bibliographystyle{plain}


\appendix

\section{Details from Section \ref{sec:experiments}} \label{app:experiments}

In this section, we give details from the experiments in Section \ref{sec:experiments}.
%
We also note that although we ran AutoML techniques across three different CCAI tasks, our experiments should not be seen as a comprehensive evaluation of AutoML methods on CCAI tasks. In particular, our experiments come with the limitations that only one trial was run per experiment (due to the a single run taking up to 50 GPU hours) and although we made reasonable choices for the AutoML methods (based on popularity) and hyperparameter ranges (based on default values), we did not run an exhaustive search across AutoML methods and hyperparameters.

Furthermore, we explicitly aimed to test AutoML performance \emph{out-of-the-box}, and we therefore did not spend the time to carefully design tailored search spaces to the tasks at hand, which would be non-trivial (e.g., see \citep{roberts2021rethinking}). However, we discuss this as a very promising area for future work in Sections \ref{subsec:discussion} and \ref{sec:conclusion}.

\subsection{ClimART} \label{app:climart}

First, we give additional details about ClimART and the corresponding experiments.

\paragraph{ClimART.}
The ClimART dataset \citep{cachay2021climart} consists of data that is simulated from CanESM5 \citep{swart2019canadian}.
This dataset takes global snapshots of the atmosphere split into a $128\times 64$ latitude-longitude grid, every 205 hours from 1979 to 2014. Each datapoint is a ``column'' of the atmosphere at a specific time, with measurements of temperature, water vapor, and aerosols taken at 49 different heights.
Each column also has global properties, such as optical and geographical information.
Prior work has tested MLPs, CNNs, GNNs, and GCNs as baselines \citep{cachay2021climart}.

\subsubsection{SMAC3 details}
We use data from years 1990, 1999, and 2003 for training, and data from 2005 for validation, to match the setting of the benchmark experiments in the original ClimART paper \citep{cachay2021climart}.
Each model is trained for 5 epochs and validated.
Hyperparameters for each model are set according to the original configurations provided by ClimART authors. 
We use a hyperparameter search space as follows: 
\begin{itemize}
    \item Unif$\{$MLP, CNN, GNN, GCN, L-GCN$\}$
    \item $\log_{10}{\text{(learning rate)}}$: Unif$[-5, -1]$  
    \item $\log_{10}{\text{(weight decay)}}$: Unif$[-7, -4]$
\end{itemize}

\subsubsection{Optuna Details}
We use a hyperparameter search space as follows: 
\begin{itemize}
    \item $\log_{10}{\text{(learning rate)}}$: Unif$[-5, -1]$  
    \item $\log_{10}{\text{(weight decay)}}$: Unif$[-7, -4]$
    \item $\text{dropout}$: Unif$[0.0, 0.8]$
    \item $\text{batch size}$: $2\text{**}{\text{int}(\text{Unif}[7.0, 9.0])}$
\end{itemize}

\begin{table}[ht]
\begin{tabular}{l|l|l|l|l|l}
\hline
                                                               & Learning rate & Weight decay & Dropout & Batch size & Test RMSE \\ \hline
\begin{tabular}[c]{@{}l@{}}36 trials/\\ 20 epochs\end{tabular} & 1.43e-4       & 2.14e-5      & 0.0     & 256        & 1.538     \\ \hline
\begin{tabular}[c]{@{}l@{}}24 trials/\\ 10 epochs\end{tabular} & 4.12e-4       & 1.96e-5      & 0.001   & 256        & 2.344     \\ \hline
original                                                       & 2e-4          & 1e-6         & 0.0     & 128        & 1.829    
\end{tabular}
\caption{Searched hyperparameters and performance comparison with original configuration. \label{tab:climart-optuna}}

\end{table}

We ran Optuna by training each architecture to 10 epochs during the search, and training each architecture to 20 epochs during the search.
The best model according to validation accuracy is fully trained to 100 epochs and then the test accuracy is compared to the original (default) model (also fully trained to 100 epochs).

\subsection{SDWPF} \label{app:sdwpf}

Next, we give details of the SDWPF dataset and experiments.

\paragraph{SDWPF.}
The SDWPF (Spatial Dynamic Wind Power Forecasting) dataset was recently featured in a KDD Cup 2022 competition that included 2490 participants \citep{zhou2022sdwpf}.%
\footnote{\url{https://aistudio.baidu.com/aistudio/competition/detail/152}}
This is by far the largest wind power forecasting dataset, consisting of data from 134 wind turbines across 12 months, with data sampled every 10 minutes.
The features consist of external features such as wind speed, wind direction, temperature, and turbine features such as nacelle direction, pitch angle of the blades, operating status, relative location, and elevation.
The problem is to predict the generated power for all 134 turbines every 10 minutes in a 48 hour time window.

We ran hyperparameter optimization over the BERT-based model with batch size and learning rate using Optuna.
Due to computational constraints, we ran the search over $25\%$ of the data and then trained the best model according to the validation set with the whole data. 

We use a hyperparameter search space as follows: 
\begin{itemize}
    \item $\log_{10}{\text{(learning rate)}}$: Unif$[-7, -1]$  
    \item $\text{batch size}$: $2\text{**}{\text{int}(\text{Unif}[5.0, 10.0])}$
    \item $\text{Feed Forward Network dropout}$: Unif$[0.0, 0.5]$
    \item $\text{Attention dropout}$: Unif$[0.0, 0.5]$
\end{itemize}

\begin{table}[h]
\begin{tabular}{l|l|l|l|l|l|l}
\hline
& Learning rate & Batch size & Test Score \\ \hline
\begin{tabular}[c]{@{}l@{}}70 trials/\\ $50\%$ data\end{tabular} 
&     4.7e-3    &     512        & -45.329\\ \hline
original                                                       
&     5e-3      &     1024       &  -45.246 
\end{tabular}
\caption{Searched hyperparameters via HPO and performance comparison with original configuration on SDWPF. \label{tab:sdwpf-optuna}}
\end{table}
\newpage


Next, we ran neural architecture search on the same BERT-based architecture, using 50\% of the data. The search space is as follows:
\begin{itemize}
    \item No. of BERT Blocks \{1,2,4\}
    \item No. of heads in attention model \{1,2,4\}
    \item Attention dropout inside BERT block Unif[0.0,0.4]
    \item Feed Forward Network dropout inside BERT block Unif[0.0,0.4]
    \item Filter sizes inside BERT Block \{8,16,32,64,128\}
\end{itemize}

\begin{table}[h]
\begin{tabular}{l|l|l|l|l|l|l}
\hline
& num blocks & num heads & attention dropout & ffn dropout & Test Score \\ \hline
\begin{tabular}[c]{@{}l@{}}40 trials/\\ $50\%$ data\end{tabular} 
&     1    &     1        & 0.224 & 0.097 & -45.178 \\ \hline
original                                                       
&     1     &     1       & 0.0 & 0.0 & -45.246
\end{tabular}
\caption{Searched model parameters via NAS and performance comparison with the original configuration on SDWPF. \label{tab:sdwpf-optuna-nas}}
\end{table}


Finally, we ran HPO on the GRU+LGBoost algorithm. We used the following hyperparameter search space:
\begin{itemize}
    \item No. of numeric embedding dimension: int(Unif[32,64])
    \item No. of time embedding dimension: int(Unif[4,8])
    \item No. of ID embedding dimension: int(Unif[4,8])
    \item No of GRU hidden units: int(Unif[32,64])
    \item $log_{10}$ (Learning rate):  Unif([-6,-2])
\end{itemize}
And the hyperparameter search space for the LGBoost model is as follows:
\begin{itemize}
    \item No. of leaves: int(Unif[2,128])
    \item Bagging frequency: int(Unif[1,7])
    \item Bagging fraction: Unif[0.4,1]
    \item Feature fraction: Unif[0.4,1]
    \item Learning rate : Unif[0.001,0.7]
\end{itemize}

\begin{table}[h]
\begin{tabular}{l|l|l|l|l|l|l}
\hline
& num\_sz & time\_sz & id\_sz & hiden & GRU\_lr & Test Score \\ \hline
\begin{tabular}[c]{@{}l@{}}20 GRU  trials +\\ 50 LGBoost trials\end{tabular} 
&     51    &     4        & 4 & 64 & 0.009538 & -45.074 \\ \hline
original                                                       
&     51     &     4       & 4 & 64 & 0.009538 & -45.074
\end{tabular}
\caption{Searched GRU model parameters and performance comparison with original configuration on SDWPF. \label{tab:sdwpf-optuna-gru}}
\end{table}
\begin{table}[ht]
\begin{tabular}{l|l|l|l|l|l|l}
\hline
& num\_lv & bag\_freq & bag\_frac & feat\_frac & LGBoost\_lr & Test Score \\ \hline
\begin{tabular}[c]{@{}l@{}}20 GRU  trials +\\ 50 LGBoost trials\end{tabular} 
&     128    &     5        & 0.998798 & 0.428377 & 0.00342 & -45.074 \\ \hline
original                    
&     128    &     5        & 0.998798 & 0.428377 & 0.00342 & -45.074 
\end{tabular}
\caption{Searched LGBoost model parameters for a sample LightGBM model and performance comparison with original configuration on SDWPF. \label{tab:sdwpf-optuna-lgb}}
\end{table}

\subsection{OC20} \label{app:oc20}

Finally, we give the details of the OC20 dataset and experiments.

\paragraph{OC20.}
The Open Catalyst 2020 (OC20) dataset \citep{chanussot2021open} was featured in a NeurIPS 2021 competition \citep{das2022open}.
Each datapoint is one reaction, where the features consist of the initial starting positions of the atoms, and the label consists of the energy needed to drive the reaction.
There are over 100 million examples in total in the original dataset. 
In our experiments, we use a down-sampled version of the OC20 IS2RE task with 
10\,000 examples where we report test accuracy on out-of-domain adsorbates. 

\subsubsection{Optuna Details}
We use the following hyperparameter search space:
\begin{itemize}
    \item $\log_{10}{\text{(learning rate)}}$: Unif$[-5, -3]$  
    \item $\log_{10}{\text{(warm-up steps)}}$: Unif$[0, 4]$  
    \item layers: Unif$[1, 12]$  
    \item attention heads: Unif$[\{6, 12, 24, 32, 48\}]$  
    \item blocks: Unif$[1, 4]$  
\end{itemize}

\begin{table}[ht]
\begin{tabular}{l|l|l|l|l|l|l}
\hline
& Learning rate & Warm up steps & Layers & Attention heads & Blocks & Test MAE \\ \hline
\begin{tabular}[c]{@{}l@{}}36 trials/\\ 4 epochs\end{tabular} 
&     2.9e-4    &     133    & 9     & 32        & 1     & 0.396 \\ \hline
original                                                       
&     3e-4      &     100    & 12    & 48        & 4     & 0.399
\end{tabular}
\caption{Searched hyperparameters and performance comparison with original configuration on OC20. \label{tab:oc20-optuna}}
\end{table}


\end{document}